\pdfoutput=1

\documentclass[11pt]{article}

\usepackage[]{EMNLP2022}

\usepackage{times}
\usepackage{latexsym}
\usepackage{amsmath}
\usepackage{multirow}
\usepackage{graphicx}
\usepackage{amsfonts}
\usepackage[T1]{fontenc}

\usepackage[utf8]{inputenc}

\usepackage{microtype}

\usepackage{inconsolata}

\usepackage{booktabs}
\usepackage{amsfonts}

\newcommand{\FullModelName}{mixture of short-channel distillers}
\newcommand{\ModelName}{MSD}

\title{\textsc{Wider \& Closer}: Mixture of Short-channel Distillers \\ for Zero-shot Cross-lingual Named Entity Recognition}

\author{
Jun-Yu Ma$^1$\thanks{\hspace{1.5mm}Equal contribution.}, Beiduo Chen$^1$\footnotemark[1], Jia-Chen Gu$^1$, Zhen-Hua Ling$^1$, Wu Guo$^1$\thanks{\hspace{1.5mm}Corresponding author.} \\
{\bf Quan Liu$^{2,3}$, Zhigang Chen$^4$, Cong Liu$^{1,3}$} \\
  $^1$National Engineering Research Center of Speech and Language Information Processing, \\
      University of Science and Technology of China, Hefei, China \\
  $^2$State Key Laboratory of Cognitive Intelligence ~ 
  $^3$iFLYTEK Research, Hefei, China \\
  $^4$Jilin Kexun Information Technology Co., Ltd. \\
{\tt \{mjy1999,beiduo\}@mail.ustc.edu.cn}, {\tt \{gujc,zhling,guowu\}@ustc.edu.cn}, \\ {\tt \{quanliu,zgchen,congliu2\}@iflytek.com}
}

\begin{document}
\maketitle
\begin{abstract}
Zero-shot cross-lingual named entity recognition (NER) aims at transferring knowledge from annotated and rich-resource data in source languages to unlabeled and lean-resource data in target languages.
Existing mainstream methods based on the teacher-student distillation framework ignore the rich and complementary information lying in the intermediate layers of  pre-trained language models, and domain-invariant information is easily lost during transfer.
In this study, a mixture of short-channel distillers (MSD) method is proposed to fully interact the rich hierarchical information in the teacher model and to transfer knowledge to the student model sufficiently and efficiently.
Concretely, a multi-channel distillation framework is designed for sufficient information transfer by aggregating multiple distillers as a mixture.
Besides, an unsupervised method adopting parallel domain adaptation is proposed to shorten the channels between the teacher and student models to preserve domain-invariant features.
Experiments on four datasets across nine languages
demonstrate that the proposed method achieves new state-of-the-art performance on zero-shot cross-lingual NER and shows great generalization and compatibility across languages and fields.

\end{abstract}

\section{Introduction}
Named entity recognition (NER) is a fundamental and important task to locate and classify named entities in a text sequence. 
Recently, deep neural networks have achieved great performance on monolingual NER in rich-resource languages with abundant labeled data~\cite{DBLP:conf/acl/YeL18,DBLP:conf/emnlp/JiaSYZ20,DBLP:conf/semeval/ChenMQGLL22}.
However, it is too expensive to annotate a large amount of data in low-resource languages for supervised NER training.
This issue drives research on cross-lingual NER, which utilizes the rich-resource annotated data in source languages to alleviate the scarcity of unlabeled lean-resource data in target languages.
In this paper, following \citet{DBLP:conf/acl/WuLKLH20}, we focus on the extremely low-resource setting, i.e., zero-shot cross-lingual NER, where labeled data is not available in target languages.

\begin{figure}[t]
\centering
\includegraphics[width=0.48\textwidth]{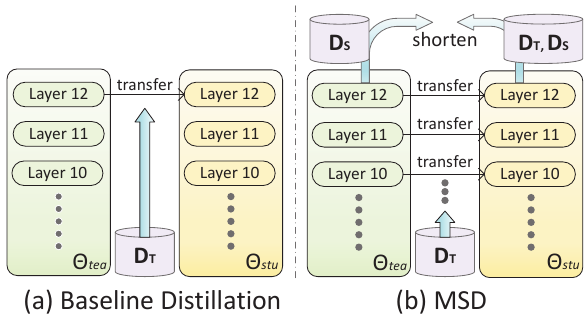}
\caption{Comparison between the previous mainstream method and the proposed MSD.
(a) \textbf{Baseline Distillation} is the teacher-student distillation framework. The teacher model is utilized to predict the soft labels of unlabeled target language data, which are further utilized to distill a student model. 
(b) \textbf{MSD} constructs a dozen of channels and shortens the transmission route between the teacher and student models to transfer NER knowledge.
\(\Theta_{tea}\) / \(\Theta_{stu}\): teacher / student models;
$\textbf{D}_\textbf{S}$ / $\textbf{D}_\textbf{T}$: unlabeled source / target language data.
}  \label{fig1}
\vspace{-4mm}
\end{figure}

The most popular approaches for zero-shot cross-lingual NER are based on distillation~\cite{DBLP:conf/acl/WuLKLH20,DBLP:conf/ijcai/WuLKHL20,DBLP:conf/acl/ChenJW0G20,DBLP:conf/kdd/LiangGPSZZJ21}. 
They employed a supervisedly trained teacher model to predict the soft labels of target languages, and then utilized the soft labels to distill a student model, which was first exploited in \citet{DBLP:conf/acl/WuLKLH20}.
Besides, domain-invariant features have been proven effective for distillation~\cite{DBLP:conf/ijcnn/Nguyen-MeidineG20,DBLP:conf/emnlp/HuWZGCS19}.
\citet{DBLP:conf/acl/ChenJW0G20} proposed to alleviate the representation discrepancy between languages in the teacher model to exploit language-independent features, which were further distilled to the student model.
It is worth noting that distillation-based methods assisted with auxiliary tasks have become the mainstream paradigm due to their robustness and scalability, achieving good performance on zero-shot cross-lingual NER~\cite{DBLP:conf/acl/LiHGCQZ22}.

However, these methods always ignored the rich and complementary information lying in the intermediate layers of multilingual BERT (mBERT) \cite{DBLP:conf/naacl/DevlinCLT19}. 
\citet{DBLP:conf/acl/PiresSG19} and \citet{DBLP:conf/eacl/MullerESS21} have verified that the upper layers of mBERT are more task-specific and not as important as the lower ones in terms of cross-language transfer.
But recent studies just adopted the last layer of mBERT for distillation \cite{DBLP:conf/ijcai/WuLKHL20,DBLP:conf/acl/LiHGCQZ22}, while neglected the explicit knowledge transfer of the lower layers.
Besides, domain-invariant features in the teacher model were first exploited, which were then transferred to the student model via distillation~\cite{DBLP:conf/acl/ChenJW0G20}.
However, due to the limitation of transfer learning,
it is difficult to fully retain the domain-invariant features to the student model.
Furthermore, auxiliary tasks to assist distillation usually require the operations of translation or data sifting \cite{DBLP:conf/ijcai/WuLKHL20}, resulting in huge pre-processing costs.
The ensemble strategies to generate high-quality soft labels or augmented data also require oceans of model parameters and a large number of computational resources.

On account of the above issues, a mixture of short-channel distillers (MSD) method is proposed in this paper to transfer cross-lingual NER knowledge sufficiently and efficiently.
On the one hand, a multi-channel distillation framework is designed to let the hierarchical information in the teacher model fully interact with each other, and transfer more complementary information to the student model.
Specifically, the teacher model is first trained on the annotated source data with each layer being directly supervised by the labels.
Then, each layer of the teacher model is tasked to predict the soft labels of the unlabeled target data.
Correspondingly, the layers of the student model are distilled by leveraging the mixture sets of soft labels from the teacher model, constructing multiple information transmission channels for a ``\emph{wider}'' bridge between the teacher and student models.
On the other hand, an unsupervised auxiliary task of parallel domain adaptation is proposed to explicitly transfer domain information.
During every batch of distillation, as Figure~\ref{fig1} depicts, unlabeled target data is fed into the student model, while unlabeled source data is fed into both the teacher and student models.

The representation discrepancy between the outputs of the teacher and student source language, together with that between the outputs of the teacher source and student target languages is minimized to preserve the cross-model and cross-language domain information respectively.
In this way, the domain information can be preserved across models and languages, so that the domains of the teacher and student models can be effectively pulled ``\emph{closer}''.

Experiments on four datasets across nine languages are conducted to evaluate the effectiveness of the proposed MSD.
The results show that our method achieves new state-of-the-art performance on all datasets.

In summary, our contributions are as follows:
(1) A multi-channel framework is proposed to leverage the rich, hierarchical and complementary information contained in the teacher model, and to interactively transfer cross-lingual NER knowledge to the student model.
(2) An unsupervised auxiliary method is designed to explicitly constrain the discrepancy of teacher/student domains without utilizing any external resources.
(3) Experiments on four datasets across nine languages verify the effectiveness and generalization ability of MSD.

\section{Related Work}

\paragraph{Zero-shot Cross-lingual NER}
Existing methods on zero-shot cross-lingual NER are mainly separated into three categories: translation-based, feature-based, and distillation-based.
Translation-based methods generate pseudo labels for the target language data from the labeled source language data.
\citet{DBLP:conf/emnlp/JainPL19} projected labels from the source language into the target language by using entity projection information.
\citet{DBLP:conf/emnlp/XieYNSC18} and \citet{DBLP:conf/ijcai/WuLKHL20} translated the annotated source language data to the target language word-by-word.
Feature-based methods use the labeled source language data to train the language model for a language-independent representation, such as Wikifier features~\cite{DBLP:conf/conll/TsaiMR16}, aligned word representations~\cite{DBLP:conf/emnlp/WuD19}, and adversarial learning encodings~\cite{DBLP:conf/emnlp/KeungLB19}.
Distillation-based methods are effective in cross-lingual NER by transferring knowledge from a teacher model to a student model~\cite{DBLP:journals/corr/HintonVD15}.
The teacher model is first trained on the labeled source language data.
Then the student model is trained on soft labels of the target language data predicted by the teacher model.
\citet{DBLP:conf/acl/WuLKLH20} trained several teacher models to generate averaged soft labels for the student model.
\citet{DBLP:conf/kdd/LiangGPSZZJ21} proposed a reinforced knowledge distillation framework to selectively transfer
useful information.

\paragraph{Domain Adaption}
Label sparsity causes domain shift~\cite{DBLP:journals/ml/Ben-DavidBCKPV10} in zero-shot cross-lingual NER.
The strategy of cross-domain transfer~\cite{DBLP:conf/ijcai/QinN0C20,DBLP:conf/emnlp/ZhangXYLZ21,DBLP:conf/emnlp/HuangAPC21} is widely adopted.
Existing methods mitigate the discrepancies of sentence patterns between the source and target domains, mainly including multi-level adaptation layers~\cite{DBLP:conf/emnlp/LinL18}, tensor decomposition~\cite{DBLP:conf/acl/JiaXZ19}, multi-task learning~\cite{DBLP:conf/rep4nlp/LiuWF20} and word alignment~\cite{DBLP:conf/cikm/LeeLLCH21}.
However, these methods require sufficient labeled data, in contrast to zero-shot scenarios.

Generally speaking, previous studies on zero-shot cross-lingual NER only leverage the last layer of the teacher model. 
Besides, the existing NER domain adaptation strategies only constrain the domain-invariant information within the teacher model and transfer them to the student model implicitly. 
To the best of our knowledge, this paper makes the first
attempt to let the rich and hierarchical information in the teacher model fully interact with each other, and further transfer the  domain information to the student model explicitly.

\section{Preliminary}

\subsection{Problem Definition}
NER is typically formulated as a sequence labeling task. 
Denote one sentence as \(\boldsymbol{x}=\{x_i\}_{i=1}^{L}\) with its labels \(\boldsymbol{y}=\{y_i\}_{i=1}^{L}\), where \(y_i\) denotes the label of its corresponding word \(x_i\) and \emph{L} denotes the length of the sentence. 
An NER model generates a sequence of predictions \(\boldsymbol{\bar{y}}=\{\bar{y}_i\}_{i=1}^{L}\), where \(\bar{y}_i\) denotes the label of \(x_i\) annotated by the model.
The labeled data \(\mathcal{D}_{\text {train }}^{S}=\{(\boldsymbol{x}, \boldsymbol{y})\}\) is available for the source language, while the unlabeled \(\mathcal{D}_{\text {train}}^{T}=\{\boldsymbol{x}\}\) and labeled \(\mathcal{D}_{\text {test}}^{T}=\{(\boldsymbol{x}, \boldsymbol{y})\}\) are available for target languages.
Formally, zero-shot cross-lingual NER aims at achieving good performance on \(\mathcal{D}_{\text {test}}^{T}\) by leveraging both \(\mathcal{D}_{\text {train}}^{S}\) and \(\mathcal{D}_{\text {train}}^{T}\).

\subsection{Basic Model} \label{base model}
The basic model for cross-lingual NER in this paper consists of  a semantic
encoder and a classifier.
The encoder \(\boldsymbol{{f}_{\theta}}\) is used to learn and generate the contextual representations of input sentences.

Following \citet{DBLP:conf/emnlp/WuD19}, the widely-used multilingual pre-trained language model, mBERT, is utilized as the encoder to extract semantic representations.
A softmax classification layer is appended to calculate the probability.
Finally, the basic model is formulated as follows:
\begin{align}
    \boldsymbol{H} &= \boldsymbol{{f}_{\theta}}(\boldsymbol{x}), \\
    \boldsymbol{p}\left(x_{i} ; \Theta\right) &= \operatorname{softmax}\left(\boldsymbol{W} \cdot \boldsymbol{h}_{i}+\boldsymbol{b}\right),
\end{align}
where \(\boldsymbol{H}=\{\boldsymbol{h_i}\}_{i=1}^{L}\) and \(\boldsymbol{h_i}\) is the representation of \(x_i\).
$\boldsymbol{p}(x_{i} ; \Theta) \in    \mathbb{R}^{|C|}$ with \emph{C} being a set of entity labels, and  $\Theta=\left\{\boldsymbol{f_{\theta}}, \boldsymbol{W}, \boldsymbol{b}\right\}$  denotes all the parameters to be learned. 

\subsection{Maximum Mean Discrepancy (MMD)} \label{mmd}
MMD~\cite{DBLP:conf/icml/LongC0J15} is a nonparametric test statistic to measure the distance between the distributions of two different random variables \((P_s,P_t)\).
MMD is defined in particular function spaces as follows:

\begin{small}
    \begin{equation}
        \begin{aligned}
            \operatorname{MMD}(\mathcal{F}, p_s, p_t)=\sup_{f \in \mathcal{F}}\left(\mathbb{E}_{x \sim p_s}[f(x)]-\mathbb{E}_{y \sim p_t}[f(y)]\right),
        \end{aligned}
    \end{equation}
\end{small}

\noindent
where \(\mathcal{F}\) is  the unit ball in a universal Reproducing Kernel Hilbert Space (RKHS) denoted by \(\mathcal{H}\).
An important property of MMD is that \(\operatorname{MMD}(\mathcal{F}, p_s, p_t)=0\) if and only if \(P_s=P_t\). 
Given the source and target sample sets \(S=\{s_i\}_{i=1}^{M}\) and \(T=\{t_j\}_{j=1}^{N}\) respectively, where \(s_i\) or \(t_j\) denotes a sample of the set, the empirical estimation of MMD can be defined as:

\begin{small}
    \begin{equation}
        \begin{aligned}
            \operatorname{MMD}(S, T)=\left\|\frac{1}{M} \sum_{i=1}^{M} \phi\left(s_{i}\right)-\frac{1}{N} \sum_{j=1}^{N} \phi\left(t_{j}\right)\right\|_{\mathcal{H}},
        \end{aligned}
    \end{equation}
\end{small}

\noindent
where \(\phi(\cdot): \mathcal{X} \rightarrow \mathcal{H}\) is a nonlinear mapping.

In cross-lingual NER, the squared formulation of MMD between the representations (\(\boldsymbol{h^s}\) or \(\boldsymbol{h^t}\)) of the two sets is usually calculated as:

\begin{small}
\begin{equation}
    \begin{aligned}
        &\operatorname{MMD}^{2}(S,T)=\frac{1}{\left(M\right)^{2}} \sum_{i, j=1}^{M} G\left(\boldsymbol{h_{i}^{s}}, \boldsymbol{h_{j}^{s}}\right)+ \\
        &\frac{1}{\left(N\right)^{2}} \sum_{i, j=1}^{N} G\left(\boldsymbol{h_{i}^{t}}, \boldsymbol{h_{j}^{t}}\right)-\frac{2}{M \times N} \sum_{i, j=1}^{M \times N} G\left(\boldsymbol{h_{i}^{s}}, \boldsymbol{h_{j}^{t}}\right),
    \end{aligned}
\end{equation}
\end{small}

\noindent
where \(G\) is a Gaussian kernel in this paper.

\begin{figure*}[h]
\centering
\includegraphics[width=0.95\textwidth]{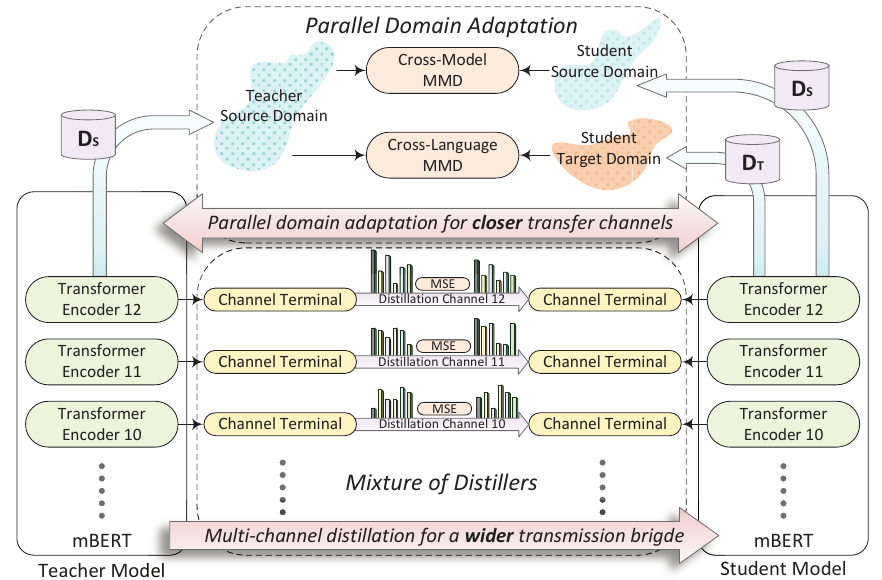}
\caption{The overall structure of the proposed MSD.}
\label{fig3}
\end{figure*}

\section{Methodology}

In this section, we present the detailed framework of the proposed \FullModelName{} (\ModelName{}).
On the one hand, the mixture of distillers module is introduced.
Specifically, multiple channels are built between corresponding layers of the teacher and student's encoders. Then a mixture of weights is employed to control the broadened information transmission route.
On the other hand, parallel domain adaptation is conducted to explicitly transfer domain information between the teacher and student models during distillation.

\subsection{Mixture of Distillers} \label{Mixture of distillers}
Previous studies have verified the importance of lower layers for cross-language transfer~\cite{DBLP:conf/eacl/MullerESS21}.
Our pilot experiments also further illustrate that lower layers of mBERT are critical for NER, which are elaborated in Appendix~\ref{prior}.
To this end, we propose the mixture of distillers framework that fully transfers the complementary information to the student model to equip it with stronger cross-language NER ability.
To establish multiple information transmission channels, each layer of the pre-trained mBERT is appended with a classifier. 
Denote each of these classifiers as a ``\emph{channel terminal}'', as shown in Figure~\ref{fig3}.
Given a sentence \(\boldsymbol{x}\) of length \emph{L} with labels \(\boldsymbol{y}\) from source language data \(\mathcal{D}_{\text {train }}^{S}\), these could be described as:
\begin{align}
    \boldsymbol{H}^{m} &= \boldsymbol{{f}_{\theta}}^{m}(\boldsymbol{x}), \\
    \boldsymbol{p}^{m}\left(x_{i} ; \Theta\right) &=\operatorname{softmax}\left(\boldsymbol{W}^{m} \cdot \boldsymbol{h}_{i}^{m}+\boldsymbol{b}^{m}\right),
\end{align}
where \(\boldsymbol{H^{m}}\) is the sentence representation from the \emph{m}-th layer of mBERT and \(\boldsymbol{p}^{m}\left(x_{i} ;
\Theta\right)\) is the probability distribution generated from the corresponding channel terminal.

At the training stage for the teacher,
the language model along with several channel terminals are jointly trained on the labeled source language data. Specifically, the channel terminal of the last layer is employed as the main distiller and the others are employed as the auxiliary ones in charge of providing complementary information. 
Following \citet{DBLP:conf/acl/WuLKLH20}, the embedding layer and the bottom three layers of mBERT in the teacher and student models are frozen.
So only the top nine layers of the teacher are optimized as:
\begin{align}
    \mathcal{L}_{\text {main }}&=\frac{1}{L}\sum_{i=1}^{L} \mathcal{L}_{\text {CE }}\left(\boldsymbol{p}^{12}\left(x_{i} ; \Theta\right), y_{i}\right), \\
    \mathcal{L}_{\text {aux }}&=\frac{1}{L}\sum_{i=1}^{L}\sum_{m=4}^{11} \lambda_{m} \mathcal{L}_{\text {CE }}\left(\boldsymbol{p}^{m}\left(x_{i} ; \Theta\right), y_{i}\right),
\end{align}
where \(\lambda_{m}\in\mathbb{R}\) is a trainable parameter representing the contribution degree of the \emph{m}-th layer\footnote{We did study semantic-wise weights by projecting the \texttt{[CLS]} token embeddings to a set of trainable parameters, but no further improvement could be achieved.} and $\mathcal{L}_{\text {CE}}$ is cross entropy loss.
The final loss for the teacher model is denoted as:
\begin{equation} \label{equation10}
    \begin{aligned}
        \mathcal{L}_{\text{tea}}= \mathcal{L}_{\text {main }}+\alpha \mathcal{L}_{\text {aux }},
    \end{aligned}
\end{equation}
where \(\alpha\) is a manually set hyperparameter that regulates the contribution of the auxiliary layers.

For the following knowledge distillation, a student model \(\Theta_{stu}\) is distilled based on the unlabeled target language data \(\mathcal{D}_{\text {train}}^{T}\).
In this paper, the student model has the same structure as the teacher model\footnote{Our method can also be extended to the framework that the teacher and student models are asymmetrical by designing a mapping function as that in \citet{DBLP:conf/emnlp/JiaoYSJCL0L20}, which will be a part of our future work.}.
Firstly, \(\mathcal{D}_{\text {train}}^{T}\) is fed into the teacher model \(\Theta_{tea}\) to obtain its soft labels derived from all the appended channel terminals.
Then, as shown in Figure~\ref{fig3}, each layer of the student model can be trained along with the student channel terminals using the mixture of soft labels generated from the corresponding layer of the teacher model.
Given a sentence \(\boldsymbol{x^{\prime}}\) of length \emph{L} from \(\mathcal{D}_{\text {train}}^{T}\),
the distillation loss of the \emph{m}-th layer is as follows:
\begin{equation}
    \begin{small}
        \begin{aligned}
            \mathcal{L}_{\text {m }}^{KD}=\frac{1}{L}\sum_{i=1}^{L}{\text {MSE }}\left(\boldsymbol{p}^{m}\left(x_{i}^{\prime} ; \Theta_{tea}\right), \boldsymbol{p}^{m}\left(x_{i}^{\prime} ; \Theta_{stu}\right)\right).
        \end{aligned}
    \end{small}
\end{equation}
Following Eq.~(\ref{equation10}), the loss for the multi-channel distillation is as follows:

\begin{equation} \label{equation12}
    \begin{aligned}
        \mathcal{L}_{\text{stu}} &=\mathcal{L}_{\text {main }}^{KD}+\beta \mathcal{L}_{\text {aux }}^{KD} \\
        &= \mathcal{L}_{\text {12}}^{KD} + \sum_{m=4}^{11} \lambda_{m}^{\prime} \mathcal{L}_{m}^{KD},
    \end{aligned}
\end{equation}
\noindent
where \(\lambda_{m}^{\prime}\) and \(\beta\) have the same effect on the student model as \(\lambda_{m}\) and \(\alpha\) do on the teacher model.

\subsection{Parallel Domain Adaptation}  \label{short channel}
As aforementioned, the teacher model is trained with hard-labeled source data, but the student model is trained with soft-labeled target data.
Thus, training in different manners and languages leads to a huge discrepancy between the domains of the teacher and student models, which causes the loss of domain information during distillation and decreases the transfer efficiency.

In this section, we aim to explicitly transfer domain information to provide a closer route for distillation. 
The parallel domain adaptation method based on MMD is proposed to preserve domain information between the teacher and student models at sentence-level. 
As Figure~\ref{fig3} depicts, the cross-model and cross-language MMD losses are proposed to minimize the cross-model and cross-language discrepancies respectively, which are denoted as \(\mathcal{L}_{MMD}^{M}\) and \(\mathcal{L}_{MMD}^{L}\).
During distillation, the soft labels \(D_{train}^{S_{tea}}\) and \(D_{train}^{S_{stu}}\) are obtained by applying the teacher and student models to the source language data respectively.
The \(\mathcal{L}_{MMD}^{M}\) could be formulated as:
\begin{equation}
    \begin{small}
        \begin{aligned}
            {\mathcal{L}_{\text{MMD}}^{M}}(D_{train}^{S_{tea}},D_{train}^{S_{stu}})=  \operatorname{MMD}^{2}(\boldsymbol{H_{cls}^{S_{tea}}}, \boldsymbol{H_{cls}^{S_{stu}}}),
        \end{aligned}
    \end{small}
\end{equation}
where $\boldsymbol{H_{cls}}$ denotes a set of \texttt{[CLS]} token embeddings $\boldsymbol{h_{cls}}$.
Meantime, the soft labels \(D_{train}^{T_{stu}}\) is obtained by applying the student model to the unlabeled target language data.
The \(\mathcal{L}_{MMD}^{L}\) is formulated as:
\begin{equation}
    \begin{small}
        \begin{aligned}
            {\mathcal{L}_{\text{MMD}}^{L}}(D_{train}^{S_{tea}},D_{train}^{T_{stu}})=  \operatorname{MMD}^{2}(\boldsymbol{H_{cls}^{S_{tea}}}, \boldsymbol{H_{cls}^{T_{stu}}}).
        \end{aligned}
    \end{small}
\end{equation}

Thus, the discrepancies between the teacher and student models, as well as between the source and target languages are both reduced during distillation, strengthening the domain adaptability of the proposed framework for efficient transfer.

The training for the final student model contains two parts: the mixture of distillers and the parallel domain adaptation.
The final loss is denoted as:
\begin{equation} \label{euqationfinal}
    \begin{aligned}
        \mathcal{L}_{\text{final}}= \mathcal{L}_{\text {stu }}+\alpha^{\prime} \mathcal{L}_{\text {MMD }}^{M}+\beta^{\prime} \mathcal{L}_{\text {MMD }}^{L},
    \end{aligned}
\end{equation}
where \(\alpha^{\prime}\) and \(\beta^{\prime}\) are the weights to balance the contributions of the parallel adaptation methods.

\section{Experiments}

In this section, the proposed MSD was evaluated on four zero-shot cross-lingual NER datasets and compared with several state-of-the-art models.
Some ablation studies were also conducted to validate the effectiveness of the proposed modules.

\subsection{Datasets}
We conducted experiments on these widely-used benchmark datasets: (1) CoNLL-2002 \cite{DBLP:conf/conll/Sang02} included Spanish and Dutch;
(2) CoNLL-2003 \cite{DBLP:conf/conll/SangM03} included English and German;
(3) WikiAnn \cite{DBLP:conf/acl/PanZMNKJ17} included English and three non-western languages (Arabic, Hindi, and Chinese);
(4) mLOWNER \cite{malmasi-etal-2022-multiconer} included four languages (English, Korean, Farsi, and Turkish).
CoNLL-2002 and CoNLL-2003 were annotated with 4 entity types:
LOC, MISC, ORG, and PER.
WikiAnn was annotated with 3 entity types: LOC, ORG, and PER.
mLOWNER was annotated with 6 entity types: LOC, ORG, PER
, CW, GRP, PROD.

In this study, the CoNLL\footnote{http://www.cnts.ua.ac.be/conll2003} and the WikiAnn\footnote{http://nlp.cs.rpi.edu/wikiann} datasets were just the same as they were initially published.
As for the mLOWNER\footnote{https://registry.opendata.aws/multiconer/} dataset, following \citet{malmasi-etal-2022-multiconer}, 10000 sentences were randomly sampled from the original test set to construct the test set used in this paper.
All datasets were annotated with the BIO entity labelling scheme and were divided into the training, development and testing sets.
Table \ref{tab-data} shows the statistics of all datasets.

Following the previous work \cite{DBLP:conf/acl/WuLKLH20}, 
English was employed as the source language in all experiments, and the other languages were employed as target languages.
Only unlabeled target language data in the training set was utilized.

\begin{table}[t]
\centering
\footnotesize
\setlength{\tabcolsep}{1.8mm}{
\begin{tabular}{ccccc}
\toprule  
Language & Type & Train & Dev & Test \\ \midrule 
\multicolumn{5}{c}{CoNLL dataset~\cite{DBLP:conf/conll/Sang02,DBLP:conf/conll/SangM03}} \\ 
\midrule 

English-en &  Sentence & 14,987 & 3,466 & 3,684 \\
(CoNLL-2003) & Entity & 23,499 & 5,942 & 5,648 \\
\hline German-de & Sentence & 12,705 & 3,068 & 3,160 \\
(CoNLL-2003) & Entity  & 11,851 & 4,833 & 3,673 \\
\hline Spanish-es & Sentence  & 8,323 & 1,915 & 1,517 \\
(CoNLL-2002) & Entity  & 18,798 & 4,351 & 3,558 \\
\hline Dutch-nl & Sentence  & 15,806 & 2,895 & 5,195 \\
(CoNLL-2002) & Entity  & 13,344 & 2,616 & 3,941 \\
\midrule

\multicolumn{5}{c}{WikiAnn dataset~\cite{DBLP:conf/acl/PanZMNKJ17}} \\ 
\midrule 
\multirow{2}{*}{English-en} &  Sentence & 20,000 & 10,000 & 10,000 \\ 
 & Entity & 27,931 & 14,146 & 13,958 \\
\hline \multirow{2}{*}{Arabic-ar} & Sentence & 20,000 & 10,000 & 10,000 \\
 & Entity  & 22,500 & 11,266 & 11,259 \\
\hline \multirow{2}{*}{Hindi-hi} & Sentence  & 5,000 & 1,000 & 1,000 \\
 & Entity  & 6,124 & 1,226 & 1,228 \\
\hline \multirow{2}{*}{Chinese-zh} & Sentence  & 20,000 & 10,000 & 10,000 \\
 & Entity  & 25,031 & 12,493 & 12,532 \\
\midrule

\multicolumn{5}{c}{mLOWNER dataset~\cite{malmasi-etal-2022-multiconer}} \\
\midrule
\multirow{2}{*}{English-en} &  Sentence & 15,300 & 800 & 10,000 \\ 
 & Entity & 23,553 & 1,230 & 15,429 \\
 \hline \multirow{2}{*}{Korean-ko} & Sentence & 15,300 & 800 & 10,000 \\
 & Entity  & 24,643 & 1,302 & 16,308 \\
\hline \multirow{2}{*}{Russian-ru} & Sentence  & 15,300 & 800 & 10,000 \\
 & Entity  & 19,840 & 1,042 & 12,941 \\
\hline \multirow{2}{*}{Turkish-tr} & Sentence  & 15,300 & 800 & 10,000 \\
 & Entity  & 23,305 & 1,245 & 15,209 \\
\bottomrule

\end{tabular}}
    \caption{The statistics of the CoNLL~\cite{DBLP:conf/conll/Sang02,DBLP:conf/conll/SangM03}, WikiAnn~\cite{DBLP:conf/acl/PanZMNKJ17} and mLOWNER~\cite{malmasi-etal-2022-multiconer} datasets.}
    \label{tab-data}
\end{table}

\subsection{Evaluation Metrics}
Following \citet{DBLP:conf/conll/Sang02}, entity-level F1-score was used as the evaluation metric.
Denote \emph{A} as the number of all entities classified by the model, \emph{B} as the number of all correct entities classified by the model, and \emph{E} as the number of all correct entities, the precision (P), recall (R), and entity-level F1-score (F1) of the model were:
\begin{align}
    \operatorname{P} = \frac{B}{A}, \ \ \operatorname{R} = \frac{B}{E}, \ \ \operatorname{F1} = \frac{2 \times P \times R}{P+R}.
\end{align}

\subsection{Baselines}
The proposed method was mainly compared with the following 
(1) distillation-based methods: \textbf{TSL} \cite{DBLP:conf/acl/WuLKLH20}, \textbf{Unitrans} \cite{DBLP:conf/ijcai/WuLKHL20}, \textbf{AdvPicker} \cite{DBLP:conf/acl/ChenJW0G20}, \textbf{RIKD} \cite{DBLP:conf/kdd/LiangGPSZZJ21}, and \textbf{MTMT} \cite{DBLP:conf/acl/LiHGCQZ22}, and 
(2) non-distillation-based methods: \textbf{Wiki}~\cite{DBLP:conf/conll/TsaiMR16}, \textbf{WS}~\cite{DBLP:conf/acl/NiDF17}, \textbf{BWET}~\cite{DBLP:conf/emnlp/XieYNSC18}, 
\textbf{Adv}~\cite{DBLP:conf/emnlp/KeungLB19}, 
\textbf{BS}~\cite{DBLP:conf/emnlp/WuD19} and \textbf{TOF}~\cite{DBLP:conf/acl/ZhangMCXZ21}.
Readers can refer to Appendix~\ref{sec-baseline} for the implementation details of the baseline models.

\subsection{Implementation Details} \label{sec-details}
All code was implemented in the PyTorch framework,\footnote{https://pytorch.org/} and is published to help replicate our results.\footnote{https://github.com/Mckysse/MSD}
All of the feature encoders mentioned in this paper employed pre-trained cased mBERT \cite{DBLP:conf/naacl/DevlinCLT19} in HuggingFace’s Transformers where the number of transformer blocks was 12, the hidden layer size was 768, and the number of self-attention heads was 12.

Some hyperparameters were empirically set following \citet{DBLP:conf/emnlp/WuD19}.
Each batch contained 32 examples, with a maximum encoding length of 128.
The dropout rate was set to 0.1, and AdamW \cite{DBLP:conf/iclr/LoshchilovH19} with WarmupLinearSchedule in the Transformers Library~\cite{wolf-etal-2020-transformers} was used as optimizer.
The parameters of the embedding layer and the bottom three layers of the mBERT used in the teacher model and the student model were frozen.

Following \citet{DBLP:conf/emnlp/KeungLB19}, the other hyperparameters were tuned on each target language dev set.
All models were trained for 10 epochs and chosen the best checkpoint with the target dev set.
For the training of teacher model, the learning rate was set to 5e-5, and the hyperparameter \(\alpha\) in Eq.~(\ref{equation10}) was set to 0.05.
For knowledge distillation, keeping the learning rate 2e-5 for the student models and the hyperparameter \(\beta\) was set to 0.05 in Eq.~(\ref{equation12}), \(\alpha^{\prime}\) and \(\beta^{\prime}\) were all set to 0.001 in Eq.~(\ref{euqationfinal}).
Furthermore, each experiment was conducted 5 times and reported the mean F1-score.

The number of parameters in a teacher or student model was about 111M.
The whole training of MSD was implemented with one GeForce RTX 3090 and consumed about 3 hours.

\begin{figure*}[h]
\centering
\includegraphics[width=0.95\textwidth]{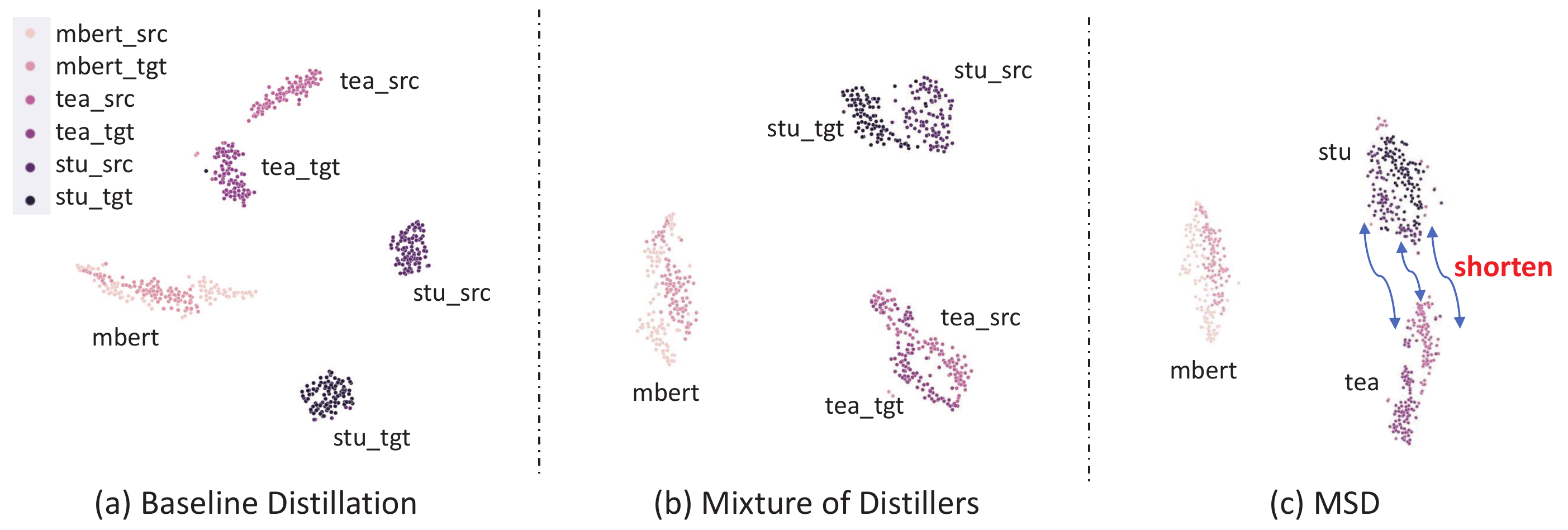}
\caption{T-SNE visualization~\cite{van2008visualizing} of semantic domains of different models by randomly sampling 100 unannotated English (source) and German (target) sentences from the training set of the CoNLL datasets~\cite{DBLP:conf/conll/Sang02,DBLP:conf/conll/SangM03}.
``\emph{tea/stu}'' refers to the teacher/student model respectively. 
``\emph{src/tgt}'' refers to the source/target data respectively.
Each point refers to the \texttt{[CLS]} representation of a sample in source/target languages. 
(a) Domains of the basic teacher-student distillation are away from each other.
(b) The distribution discrepancy within the teacher (\texttt{tea\_src}, \texttt{tea\_tgt}) or within the student (\texttt{stu\_src}, \texttt{stu\_tgt}) models is implicitly affected by the mixture of distillers.
(c) The distribution discrepancy between the teacher (\texttt{tea}) and student (\texttt{stu}) models is reduced after further performing parallel domain adaptation.
}  \label{fig2}
\end{figure*}

\subsection{Results and Comparisons}

\begin{table}[t]
\footnotesize
\setlength{\tabcolsep}{1.0mm}
\begin{tabular}{lcccc}
\toprule
\textbf{Method} & \textbf{de} & \textbf{es} & \textbf{nl} & \textbf{Avg} \\ 
\midrule

Wiki \cite{DBLP:conf/conll/TsaiMR16} &  48.12 & 60.55 & 61.56 & 56.74 \\
WS \cite{DBLP:conf/acl/NiDF17} &  58.50 & 65.10 & 65.40 & 63.00 \\
BWET \cite{DBLP:conf/emnlp/XieYNSC18} &  57.76 & 72.37 & 71.25 & 67.13 \\ 
ADV \cite{DBLP:conf/emnlp/KeungLB19} &  71.90 & 74.30 & 77.60 & 74.60 \\ 
BS \cite{DBLP:conf/emnlp/WuD19} &  69.59 & 74.96 & 77.57 & 73.57 \\
TSL \cite{DBLP:conf/acl/WuLKLH20} &  73.16 & 76.75 & 80.44 & 76.78 \\ 
Unitrans \cite{DBLP:conf/ijcai/WuLKHL20} &  74.82 & 79.31 & 82.90 & 79.01  \\ 
AdvPicker \cite{DBLP:conf/acl/ChenJW0G20} &  75.01 & 79.00 & 82.90 & 78.97 \\ 
RIKD \cite{DBLP:conf/kdd/LiangGPSZZJ21} &  75.48 & 77.84 & 82.46 & 78.59 \\ 
TOF \cite{DBLP:conf/acl/ZhangMCXZ21} &  76.57 & 80.35 & 82.79 & 79.90 \\ 
MTMT \cite{DBLP:conf/acl/LiHGCQZ22} &  76.80 & 81.82 & 83.41 & 80.67 \\ 
\midrule
\textbf{MSD} &  \textbf{77.56} & \textbf{81.92} & \textbf{85.11} & \textbf{81.53} \\ 
\midrule
MSD w/o. distillers &  75.31  & 79.34  & 83.16  & 79.27  \\ 
MSD w/o. $\mathcal{L}_{\text{MMD}}^{L}$ &  76.68  & 80.27  & 84.07  & 80.34  \\
MSD w/o. $\mathcal{L}_{\text{MMD}}^{M}$ &  77.12  & 79.81  & 84.36  & 80.43  \\ 
MSD w/o. all &  74.17  & 77.82  & 81.31  & 77.76  \\

\bottomrule

\end{tabular}
    \caption{Evaluation results (\%) of entity-level F1-score on the test set of the CoNLL datasets~\cite{DBLP:conf/conll/Sang02,DBLP:conf/conll/SangM03}. Results except ours were cited from the published literature.
    For a fair comparison, scores of the version of RIKD (mBERT) was listed.}
    \label{conll-result}
\end{table}

\begin{table}[t]
\footnotesize
\setlength{\tabcolsep}{1.3mm}
\begin{tabular}{lcccc}
\toprule  
\textbf{Method} & \textbf{ar} & \textbf{hi} & \textbf{zh} & \textbf{Avg} \\ \midrule

 BS \cite{DBLP:conf/emnlp/WuD19} &  42.30 & 67.60 & 52.90 & 54.27 \\
 TSL \cite{DBLP:conf/acl/WuLKLH20} &  43.12 & 69.54 & 48.12 & 53.59 \\ 
 RIKD \cite{DBLP:conf/kdd/LiangGPSZZJ21} &  45.96 & 70.28 & 50.40 & 55.55 \\ 
 MTMT \cite{DBLP:conf/acl/LiHGCQZ22} &  52.77 & 70.76 & 52.26 & 58.60 \\ 
\midrule
\textbf{MSD} &  \textbf{62.88} & \textbf{73.43} & \textbf{57.06} & \textbf{64.46} \\ 

\midrule
MSD w/o. distillers & 54.52 & 70.22 & 52.46 & 59.06   \\ 
MSD w/o. $\mathcal{L}_{\text{MMD}}^{L}$ & 56.93 & 71.50 & 56.68 & 61.70   \\ 
MSD w/o. $\mathcal{L}_{\text{MMD}}^{M}$ & 58.65 & 72.11 & 56.53 & 62.43       \\
MSD w/o. all & 43.17 & 68.07 & 49.25 & 53.49    \\

\bottomrule

\end{tabular}
    \caption{Evaluation results (\%) of entity-level F1-score on the test set of the WikiAnn dataset~\cite{DBLP:conf/acl/PanZMNKJ17}. Results except ours were cited from the published literature.}
    \label{wiki-result}
\end{table}

\begin{table}[h]
\footnotesize
\setlength{\tabcolsep}{1.0mm}
\begin{tabular}{lcccc}
\toprule  
\textbf{Method} & \textbf{ko} & \textbf{ru} & \textbf{tr} & \textbf{Avg} \\ \midrule 
 BS \cite{DBLP:conf/emnlp/WuD19} &  51.78 & 52.33 & 58.85 & 54.32 \\
 TSL \cite{DBLP:conf/acl/WuLKLH20} &  53.91 & 54.26 & 61.15 & 56.44 \\ 
 AdvPicker \cite{DBLP:conf/acl/ChenJW0G20} & 56.22 & 55.65 & 63.17 & 58.34 \\
\midrule
\textbf{MSD} &  \textbf{61.67} & \textbf{58.06} & \textbf{67.80} & \textbf{62.51} \\ 

\midrule
MSD w/o. distillers & 57.23 & 56.81 & 65.14 & 59.72   \\ 
MSD w/o. $\mathcal{L}_{\text{MMD}}^{L}$ & 57.88 & 57.24 	& 67.83 & 60.98 \\ 
MSD w/o. $\mathcal{L}_{\text{MMD}}^{M}$ & 59.12 & 58.08 & 67.41 & 61.53   \\
MSD w/o. all & 54.37 &	54.03 & 61.55 & 56.65 \\

\bottomrule

\end{tabular}
    \caption{Evaluation results (\%) of entity-level F1-score on the test set of the mLOWNER dataset~\cite{malmasi-etal-2022-multiconer}. Results except ours were obtained by re-implementing these baseline models with the source code provided by the original authors. 5 experiments under the same configuration were conducted for all the methods and the average results were taken as the final numbers. Numbers in bold denote that the improvement over the best performing baseline is statistically significant (t-test with \(p\)-value \textless \(0.05\)).}
    \label{mlow-result}
\end{table}

Table~\ref{conll-result}, \ref{wiki-result} and \ref{mlow-result} reported the zero-shot cross-lingual NER results of different methods on 4 datasets, containing 9 target languages.
The results show that the proposed MSD method significantly outperformed the baseline method TSL and achieved new state-of-the-art performance on all target languages.
For results on CoNLL, MSD outperformed MTMT (previous SOTA) by absolute margins of 0.76\% and 1.70\% in terms of German[de] and Dutch[nl] respectively.
As for results of non-western languages on WikiAnn and mLOWNER, MSD outperformed MTMT and AdvPicker by marked absolute margins from 2.41\% to 10.11\% for all target languages.
The results clearly demonstrated the effectiveness and generalization ability across languages and datasets of MSD.

Obviously, existing distillation-based methods were outperformed by the proposed MSD.
Specifically, translation and ensemble of teacher models for high-quality soft labels in Unitrans and AdvPicker, as well as iterative knowledge distillation in RIKD requiring huge computational resources, were not adopted in MSD any more.
Instead, the proposed MSD fully explored the rich hierarchical information in the teacher model without ensemble, and only utilized the unsupervised data without extra data-process.

Besides, AdvPicker shortened the gap between the source and target languages to derive the language-independent features in the teacher model, and then distilled the domain information to the student model implicitly.
However, the proposed MSD chose to transfer the domain-invariant information directly from the teacher to the student via the parallel domain adaptation.
The results demonstrated that the implicit domain transfer in AdvPicker is overshadowed by the explicit domain transfer in MSD.
As shown in Figure~\ref{fig2}, domain discrepancy between the teacher and student models is vividly reduced by MSD, contributing to a closer transfer route.
Further analysis of the difference between the domain transfer manners of AdvPicker and MSD was elaborated in Section~\ref{transfer manners}.

\begin{figure*}[h]
\centering
\includegraphics[width=\textwidth]{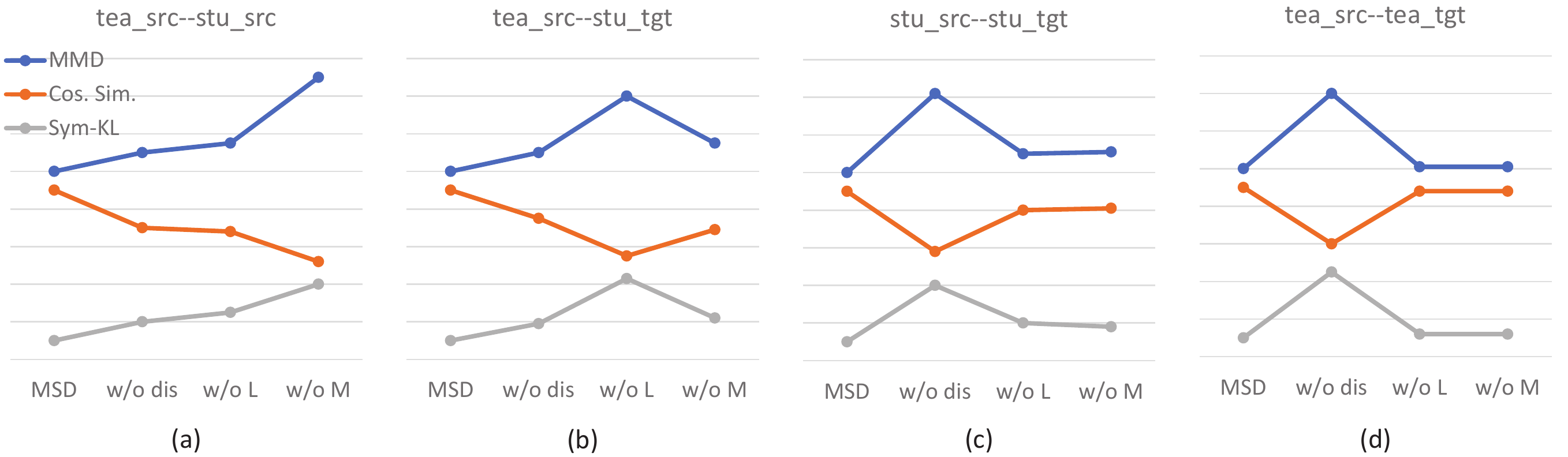}
\caption{Metrics of MMD, symmetrical KL divergence and cosine similarity on CoNLL02/03 under the ablation settings.
``tea/stu'' refers to the teacher/student model respectively. ``src/tgt'' refers to the source/target data respectively. For example, \texttt{tea\_src}\texttt{-}\texttt{-}\texttt{stu\_src} refers to the distance between the teacher's source domain and the student's source domain.
Each result was calculated from the mean center of the two domains as the same as mentioned in Section~\ref{short channel}.
[\texttt{w/o dis}]: MSD w/o. distillers; [\texttt{w/o L}]: MSD w/o. \(\mathcal{L}_{\text{MMD}}^{L}\); [\texttt{w/o L}]: MSD w/o. \(\mathcal{L}_{\text{MMD}}^{M}\).
}
\label{fig4}
\vspace{-2mm}
\end{figure*}

\subsection{Analysis} \label{sec-analysis}
\subsubsection{Ablation Study}
To validate the contributions of different components in MSD, the following variants and baselines were conducted to perform the ablation study: (1) MSD w/o. distillers, which only activated the last channel and removed others between the teacher and student models.
Besides, the two different MMD losses were still used during distillation.
(2) MSD w/o. \(\mathcal{L}_{\text{MMD}}^{L}\), which only removed the cross-language MMD loss.
In this case, the teacher and student models had multiple transmission channels but only the cross-model MMD loss was employed. 
(3) MSD w/o. \(\mathcal{L}_{\text{MMD}}^{M}\), which only removed the cross-model MMD loss correspondingly.
(4)  MSD w/o. all, which removed all the components mentioned above, and was equivalent to a baseline distillation model as TSL.

Results of the ablation experiments were shown in the bottom five lines of Table~\ref{conll-result}, ~\ref{wiki-result} and ~\ref{mlow-result} respectively.
Some in-depth analysis could be explored: (1) Compared MSD with MSD w/o. distillers, we could see that the removal of distillers caused a significant performance drop, which further demonstrated the importance of leveraging information contained in the intermediate layers of mBERT;
(2) Compared MSD with MSD w/o. \(\mathcal{L}_{\text{MMD}}^{L}\) and MSD w/o. \(\mathcal{L}_{\text{MMD}}^{M}\), the parallel domain adaptation contributed to cross-lingual NER significantly.
The results well demonstrated that explicitly transfer domain information across models and languages during distillation was reasonable and effective;
(3) The two MMD losses were correlated, as \(\mathcal{L}_{\text{MMD}}^{L}\) measured both cross-language and cross-model effects. Removal of \(\mathcal{L}_{\text{MMD}}^{L}\) caused larger performance degradation than removal of \(\mathcal{L}_{\text{MMD}}^{M}\).

The ablation study validated the effectiveness of all components.
Moreover, subtle integration of these modules achieved state-of-the-art performance.
Not only multiple channels between the teacher and student models should be established to leverage the complementary and hierarchical information of mBERT, but also these channels should be shortened for efficient transfer.

\subsubsection{Case Study on Domain Discrepancy}
To further illustrate the effectiveness of MSD, various metrics were employed to measure the distribution discrepancy between different domains: MMD~\cite{DBLP:conf/icml/LongC0J15}, symmetrical KL divergence~\cite{DBLP:conf/acl/JiangHCLGZ20,DBLP:journals/corr/abs-2004-08994}, and cosine similarity~\cite{DBLP:journals/ijprai/BromleyBBGLMSS93}.
The results in Figure~\ref{fig4} is corresponding to the ablation study.
Along with Figure~\ref{fig2}, the effects of different components could be discussed.

\romannumeral1 ) Parallel Domain Adaptation. \(\mathcal{L}_{\text{MMD}}^{M}\) pulled the source domain of the teacher and student models closer.
As shown in Figure~\ref{fig4} (a), the MMD and symmetrical KL divergence increased and the cosine similarity decreased without \(\mathcal{L}_{\text{MMD}}^{M}\).
Similar to \(\mathcal{L}_{\text{MMD}}^{M}\), \(\mathcal{L}_{\text{MMD}}^{L}\) pulled the source domain of the teacher and the target domain of the student closer.
\romannumeral2 ) Distillers. From Figure~\ref{fig4} (c) and (d), distillers made the domains within the model closer.
This effect can be seen intuitively in Figure~\ref{fig2} (b).
From  Figure~\ref{fig4} (a) and (b), the influence of distillers on the discrepancy between different models was much smaller than that of \(\mathcal{L}_{\text{MMD}}^{L}\) and \(\mathcal{L}_{\text{MMD}}^{M}\).
\romannumeral3 ) Other results. \(\mathcal{L}_{\text{MMD}}^{L}\) and \(\mathcal{L}_{\text{MMD}}^{M}\) were helpful for reducing the distance between the source and target domains of the student model, as shown in Figure~\ref{fig4} (c).
Besides, they alleviated domains discrepancy during distillation, as shown in Figure~\ref{fig2} (c).

\subsubsection{Comparison of Transfer Manners} \label{transfer manners}
To validate the effectiveness of the explicit domain transfer in MSD, an implicit domain transfer experiment was designed.
Imitating \citet{DBLP:conf/acl/ChenJW0G20}, MMD was employed to get language-independent features in the teacher model, and then a baseline distillation was conducted.
In contrast, MSD w/o. distillers actually adopted an explicit domain transfer manner.
As shown in Appendix~\ref{add}, MSD w/o. distillers outperformed methods with implicit domain transfer manners.

\section{Conclusion}
In this paper, we propose a mixture of short-channel distillers framework for zero-shot cross-lingual NER, including a multi-channel distillation framework to fully leverage the complementary and hierarchical information in the teacher model, and an unsupervised parallel domain adaptation method to effectively pull the domains between teacher and student models closer.
Experimental results show that the proposed method outperforms previous methods on four datasets across nine languages.
In the future, we will extend this method to languages where data resources are scarcer.

\section*{Limitations}
Our method has certain limitations, such as it cannot be used for target languages without any text data.
Furthermore, although the results show great performance, more efforts are required to explore the hidden impact of distillers as shown in the t-SNE graph, which will help the application of the proposed model in the future.

\section*{Acknowledgements}
We thank anonymous reviewers for their valuable comments.
This work was partially funded by the National Natural Science Foundation of China (Grant No. U1836219).

\clearpage
\appendix

\section{Appendices}

\subsection{Baseline Models} \label{sec-baseline}
We mainly compared our method with the following distillation-based methods.

\noindent
\textbf{TSL} \cite{DBLP:conf/acl/WuLKLH20} proposed a teacher-student
learning model, via using source-language models as teachers to train a student model on unlabeled data in the target language for cross-lingual NER.

\noindent
\textbf{Unitrans} \cite{DBLP:conf/ijcai/WuLKHL20}  unified both model transfer and data transfer based on their complementarity via enhanced knowledge distillation on unlabeled target-language data.

\noindent
\textbf{AdvPicker} \cite{DBLP:conf/acl/ChenJW0G20} proposed a novel approach to combine the feature-based method and pseudo labeling via language adversarial learning for cross-lingual NER.

\noindent
\textbf{RIKD} \cite{DBLP:conf/kdd/LiangGPSZZJ21} proposed a reinforced knowledge distillation framework.

\noindent
\textbf{MTMT} \cite{DBLP:conf/acl/LiHGCQZ22} proposed an unsupervised multiple-task and multiple-teacher model for cross-lingual NER.

In addition, \textbf{Wiki}~\cite{DBLP:conf/conll/TsaiMR16}, \textbf{WS}~\cite{DBLP:conf/acl/NiDF17}, \textbf{BWET}~\cite{DBLP:conf/emnlp/XieYNSC18}, 
\textbf{Adv}~\cite{DBLP:conf/emnlp/KeungLB19}, 
\textbf{BS}~\cite{DBLP:conf/emnlp/WuD19} and \textbf{TOF}~\cite{DBLP:conf/acl/ZhangMCXZ21} were non-distillation-based methods.

\subsection{Pilot Experiment} \label{prior}
Previous studies have verified the importance of lower layers for cross-language transfer~\cite{DBLP:conf/eacl/MullerESS21}.
Our pilot experiments also further illustrate that lower layers of mBERT are critical for NER as shown in Table~\ref{conll-prior}.

\begin{table}[h]
\centering
\footnotesize
\setlength{\tabcolsep}{3.0mm}
\begin{tabular}{lcccc}
\toprule  
\textbf{Method} & \textbf{en}  \\ \midrule 
 BS \cite{DBLP:conf/emnlp/WuD19} &  91.30 \\
\textbf{mixture} &  \textbf{91.49} \\
\bottomrule
\end{tabular}
\caption{Evaluation results (\%) of entity-level F1-score on the English test set of the CoNLL dataset~\cite{DBLP:conf/conll/Sang02,DBLP:conf/conll/SangM03}.
Both models were trained with the training set data of English in CoNLL.
BS was the basic model in Section~\ref{base model}.
mixture represented the teacher model described in Section~\ref{Mixture of distillers}.}
\label{conll-prior}
\end{table}

\subsection{Comparison of Transfer Manners} \label{add}

Table~\ref{add-conll}, \ref{add-wiki} and \ref{add-mlow} reported the results of different transfer manners.
Imitating  AdvPicker \cite{DBLP:conf/acl/ChenJW0G20}, TSL w. MMD was designed to get language-independent features in the teacher model, and then a baseline distillation was conducted, which was an implicit transfer manner.
MSD w/o. distillers represented the explicit transfer manner.

\begin{table}[h]
\centering
\footnotesize
\setlength{\tabcolsep}{1.3mm}
\begin{tabular}{lcccc}
\toprule  
\textbf{Method} & \textbf{de} & \textbf{es} & \textbf{nl} & \textbf{Avg} \\ 
\midrule
AdvPicker &75.01 &79.00 &82.90 &78.97 \\
TSL w. MMD & 75.02 & 78.83 & 82.73 & 78.86  \\   
\textbf{MSD w/o. distillers} & \textbf{75.31} & \textbf{79.34} & \textbf{83.16} & \textbf{79.27}   \\ 
\bottomrule
\end{tabular}
    \caption{Evaluation results (\%) of entity-level F1-score on the test set of the CoNLL datasets~\cite{DBLP:conf/conll/Sang02,DBLP:conf/conll/SangM03}.}
\label{add-conll}
\end{table}

\begin{table}[h]
\centering
\footnotesize
\setlength{\tabcolsep}{1.3mm}
\begin{tabular}{lcccc}
\toprule  
\textbf{Method} & \textbf{ar} & \textbf{hi} & \textbf{zh} & \textbf{Avg} \\ 

\midrule
AdvPicker &53.12 &69.88 &51.09 &57.69 \\
TSL w. MMD  & 53.47 & 70.09 & 50.12 & 57.89  \\   
\textbf{MSD w/o. distillers} & \textbf{54.52} & \textbf{70.22} & \textbf{52.46} & \textbf{59.06}   \\ 
\bottomrule

\end{tabular}
    \caption{Evaluation results (\%) of entity-level F1-score on the test set of the WikiAnn dataset~\cite{DBLP:conf/acl/PanZMNKJ17}.}
\label{add-wiki}
\end{table}

\begin{table}[h]
\centering
\footnotesize
\setlength{\tabcolsep}{1.3mm}
\begin{tabular}{lcccc}
\toprule  
\textbf{Method} & \textbf{ko} & \textbf{ru} & \textbf{tr} & \textbf{Avg} \\ 

\midrule
AdvPicker &56.22 &55.65 &63.17 &58.34 \\
TSL w. MMD  & 56.68 & 56.28 & 63.13 & 58.69  \\   
\textbf{MSD w/o. distillers} & \textbf{57.23} & \textbf{56.81} & \textbf{65.14} & \textbf{59.72}   \\ 

\bottomrule

\end{tabular}
    \caption{Evaluation results (\%) of entity-level F1-score on the test set of the mLOWNER dataset~\cite{malmasi-etal-2022-multiconer}.}
\label{add-mlow}
\end{table}

\end{document}